\documentclass{article} 
\usepackage{iclr2015,times}
\usepackage{url}
\usepackage[psamsfonts]{amssymb}
\usepackage{amsmath,fancybox,tikz,pgfplots,verbatim}
\usepackage{graphicx,caption,subcaption}
\usepackage{algorithm,enumitem}
\usepackage[noend]{algpseudocode}

\usetikzlibrary{arrows,calc,fit,positioning,shapes}
\pgfplotsset{compat=newest}
\pgfplotsset{plot coordinates/math parser=false}
\pgfplotsset{
  tick label style = {font=\scriptsize\sffamily},
  every axis label = {font=\footnotesize},
  legend style = {font=\tiny},
  label style = {font=\footnotesize}
}
\newlength\figurewidth
\newlength\figureheight
\newcommand{\RR}{I\!\!R} 

\def\imagetop#1{\vtop{\null\hbox{#1}}}

\DeclareMathOperator*{\argmin}{arg\,min}

\DeclareMathOperator*{\supp}{supp}

\title{Example Selection For Dictionary Learning}

\author{
Tomoki Tsuchida \& Garrison W. Cottrell \\
Department of Computer Science and Engineering\\
University of California, San Diego\\
9500 Gilman Drive, Mail Code 0404 \\
La Jolla, CA 92093-0404, USA \\
\texttt{\{ttsuchida,gary\}@ucsd.edu}
}

\iclrfinalcopy 

\begin{document}

\maketitle

\begin{abstract}
In unsupervised learning, an unbiased uniform sampling strategy is typically used, in order that the learned features faithfully encode the statistical structure of the training data.  In this work, we explore whether active example selection strategies --- algorithms that select which examples to use, based on the current estimate of the features --- can accelerate learning. Specifically, we investigate effects of heuristic and saliency-inspired selection algorithms on the dictionary learning task with sparse activations.  We show that some selection algorithms do improve the speed of learning, and we speculate on why they might work.
\end{abstract}

\section{Introduction}

The efficient coding hypothesis, proposed by \citet{Barlow:1961p1757}, posits that the goal of perceptual system is to encode the sensory signal in such a way that it is efficiently represented.  Based on this hypothesis, the past two decades have seen successful computational modeling of low-level perceptual features based on dictionary learning with sparse codes. The idea is to learn a set of dictionary elements that encode ``naturalistic'' signals efficiently; the learned dictionary might then model the features of early sensory processing. Starting with \citet{Olshausen:1996p2797}, the dictionary learning task has thus been used extensively to explain early perceptual features.  Because the objective of such a learning task is to capture the statistical structure of the observed signals faithfully and efficiently, it is an instance of unsupervised learning. As such, the dictionary learning is usually performed using \emph{unbiased} sampling: the set of data to be used for learning are sampled uniformly from the training dataset.

At the same time, the world contains an overabundance of sensory information, requiring organisms with limited processing resources to select and process only information relevant for survival \cite{Tsotsos:1990vv}.  This selection process can be expressed as perceptual action or attentional filtering mechanisms.  This might at first appear at odds with the goal of the dictionary learning task, since the selection process necessarily biases the set of observed data for the organism.  However, the converse is also true: as better (or different) features are learned over the course of learning, the mechanisms for selecting what is relevant may change, even if the selection objective stays the same.  If a dictionary learning task is to serve as a realistic algorithmic model of the feature learning process in organisms capable of attentional filtering, this mutual dependency between the dictionary learning and attentional sample selection bias must be taken into consideration.

In this work, we examine the effect of such sampling bias on the dictionary learning task.  In particular, we explore interactions between learned dictionary elements and example selection algorithms.  We investigate whether any selection algorithm can approach, or even improve upon, learning with unbiased sampling strategy.  Some of the heuristics we examine also have close relationships to models of attention, suggesting that they can be plausibly implemented by organisms evolving to effectively encode stimuli from their environment.

\section{Dictionary Learning}
\label{sec:dictionary_learning}

Assume that a training set consisting of $N$ $P$-dimensional signals ${\bf X}_{N} \triangleq \{ {\bf x}^{(i)} \}_{i=1}^N$ is generated from a $K$-element ``ground-truth'' dictionary set ${\bf A}^* = [{\bf a}_1 {\bf a}_2 \cdots {\bf a}_K]$ under the following model:

\begin{equation} \label{eqn:genmodel}
  \begin{aligned}
  {\bf x}^{(i)} &= {\bf A}^* {\bf s}^{(i)} + \epsilon^{(i)}, \\
    \{s^{(i)}_j : s^{(i)}_j > 0\} &\sim Exp(\lambda) \quad \textrm{iid},\\
  	\epsilon^{(i)} &\sim  \mathcal{N}(0, {\bf I} \sigma_\epsilon^2) \quad \textrm{iid}.
\end{aligned}
\end{equation}

Each signal column vector ${\bf x}^{(i)}$ is restricted to having exactly $k$ positive activations: ${\bf s}^{(i)} \in \mathcal{C}_s \triangleq \{ {\bf s} \in \RR_{\ge 0}^P : \|{\bf s}\|_0 = k \}$, and each dictionary element is constrained to the unit-norm: ${\bf A}^* \in \mathcal{C}_{\bf A} \triangleq \{{\bf A} : \|({\bf A})_j\|_2 = 1 \; \forall j\}$. The goal of dictionary learning is to recover ${\bf A}^*$ from ${\bf X}_N$, assuming $\lambda$ and $\sigma_\epsilon^2$ are known. To that end, we wish to calculate the maximum a posteriori estimate of ${\bf A}^*$,

\begin{equation} \label{eqn:aml}
\begin{aligned}
	\argmin_{{\bf A} \in \mathcal{C}_{\bf A}} \frac{1}{N} \sum_{i=1}^N \min_{{\bf s}^{(i)} \in \mathcal{C}_s} \left (  \frac{1}{2\sigma_\epsilon^2} \| {\bf x}^{(i)} - {\bf A} {\bf s}^{(i)} \|_2^2 + \lambda \| {\bf s}^{(i)} \|_1  \right ) .
\end{aligned}
\end{equation}

This is difficult to calculate, because ${\bf A}$ and $\{{\bf s}^{(i)}\}_{i=1}^N$ are simultaneously optimized.  One practical scheme is to fix one variable and alternately optimize the other, leading to subproblems

\begin{align}
 \label{eqn:encoding}
	{\bf {\hat S}} &= \left[ \argmin_{{\bf s}^{(i)} \in \mathcal{C}_s} \left (  \frac{1}{2\sigma_\epsilon^2} \| {\bf x}^{(i)} - {\bf {\hat A}} {\bf s}^{(i)} \|_2^2 + \lambda \|{\bf s}^{(i)}\|_1 \right ) \right]_{i=1}^N \textrm{,} \\
 \label{eqn:updating}
	{\bf {\hat A}} &= \argmin_{{\bf A} \in \mathcal{C}_{\bf A}} \frac{1}{2N} \| {\bf X}_N - {\bf A} {\bf {\hat S}} \|_F^2.
\end{align}

As in the Method of Optimal Directions (MOD)~\cite{Engan:1999kg}, this alternate optimization scheme is guaranteed to converge to a locally optimal solution for ${\bf {\hat A}}_{\textrm{MAP}}$ estimation problem~\eqref{eqn:aml}. This scheme is also attractive as an algorithmic model of low-level feature learning, since each optimization process can be related to the ``analysis'' and ``synthesis'' phases of an autoencoder network~\cite{Olshausen:1997uh}. In this paper, we henceforth refer to problems \eqref{eqn:encoding} and \eqref{eqn:updating} as \emph{encoding} and \emph{updating} stages, and their corresponding optimizers as $f_{enc}$ and $f_{upd}$.

\subsection{Encoding algorithms}

The $L^0$-constrained encoding problem \eqref{eqn:encoding} is NP-Hard~\cite{elad}, and various approximation methods have been extensively studied in the sparse coding literature. One approach is to ignore the $L^0$ constraint and solve the remaining nonnegative $L^1$-regularized least squares problem

\begin{align}
\textrm{\tt LARS}: {\hat {\bf s}}^{(i)} &= \argmin_{{\bf s} \ge 0} \left (  \frac{1}{2\sigma_\epsilon^2} \| {\bf x}^{(i)} - {\bf {\hat A}} {\bf s} \|_2^2 + \lambda' \|{\bf s} \|_1 \right ), 
\end{align}

\noindent with a larger sparsity penalty $\lambda' \triangleq \lambda P / k$ to compensate for the lack of the $L^0$ constraint. This works well in practice, since the distribution of $s_j^{(i)}$ (whose mean is $1/\lambda'$) is well approximated by $Exp(\lambda')$.  For our simulations, we use the Least Angle Regression (LARS) algorithm~\cite{lars} implemented by the SPAMS package~\cite{Mairal:2010us} to solve this.

Another approach is to greedily seek nonzero activations to minimize reconstruction errors.  The matching pursuit family of algorithms operate on this idea, and they effectively approximate the encoding model

\begin{align}
\begin{split}
\textrm{\tt OMP}: \quad {\hat {\bf s}}^{(i)} =\argmin_{{\bf s} \ge 0} \left (  \frac{1}{2\sigma_\epsilon^2} \| {\bf x}^{(i)} - {\bf {\hat A}} {\bf s} \|_2^2 \right) \\ \textrm{s.t.} \; \|{\bf s}\|_0 \le k .
\end{split}
\end{align}

This approximation ignores the $L^1$ penalty, but because nonzero activations are exponentially distributed and mostly small, this approximation is also effective.  We use the Orthogonal Matching Pursuit (OMP) algorithm~\cite{mallat1993matching}, also implemented by the SPAMS package, for this problem.

An even simpler variant of the pursuit-type algorithm is the thresholding~\cite{elad} or the $k$-Sparse algorithm~\cite{Makhzani:2013wn}. This algorithm takes the $k$ largest values of ${\bf {\hat A}}^\intercal {\bf x}^{(i)}$ and sets every other component to zero:

\begin{align}
\textrm{\tt k-Sparse}: \quad {\hat {\bf s}}^{(i)} &=\supp_k \{ {\bf {\hat A}}^\intercal {\bf x}^{(i)} \}
\end{align}

This algorithm is plausibly implemented in a feedforward phase of an autoencoder with a hidden layer that competes horizontally and picks $k$ ``winners''.  The simplicity of this algorithm is important for our purposes, because we allow the training examples to be selected \emph{after} the encoding stage, and the encoding algorithm must operate on a much larger number of examples than the updating algorithm. This view also motivated the nonnegative constraint on ${\bf s}^{(i)}$, because the activations of the hidden layers are likely to be conveyed by nonnegative firing rates.

\subsection{Dictionary update algorithm}

For the updating stage, we only consider the stochastic gradient update, another simple algorithm for learning. For the reconstruction loss $L_{rec}({\bf A}) \triangleq \frac{1}{2N} \|{\bf X}_N - {\bf A} {\bf {\hat S}}\|_F^2$, the gradient is $\nabla L_{rec} = 2 ({\bf A} {\bf {\hat S}} - {\bf X}) {\bf {\hat S}}^\intercal / N$, yielding the update rule

\begin{align}
	\label{eq:sgd}
\quad {\bf {\hat A}} \leftarrow {\bf {\hat A}} - \eta_t ({\bf {\hat A}} {\bf {\hat S}}-{\bf X}_N) {\bf {\hat S}}^\intercal / N .
\end{align}

Here, $\eta_t$ is a learning rate that decays inversely with the update epoch $t$: $\eta_t \in \Theta(1/t + c)$.  After each update, ${\bf {\hat A}}$ is projected back to ${\mathcal C}_{\bf A}$ by normalizing each column.  Given a set of training examples, this encoding and updating procedure is repeated a small number of times (10 times in our simulations).

\subsection{Activity equalization}

One practical issue with this task is that a small number of dictionary elements tend to be assigned to a large number of activations.  This produces ``the rich get richer'' effect: regularly used elements are more often used, and unused elements are left at their initial stages.  To avoid this, an activity normalization procedure takes place after the encoding stage. The idea is to modulate all activities, so that the mean activity for each element is closer to the across-element mean of the mean activities; this is done at the cost of increasing the reconstruction error.  The equalization is modulated by $\gamma$, with $\gamma = 0$ corresponding to no equalization and $\gamma = 1$ to fully egalitarian equalization (\emph{i.e.} all elements would have equal mean activities).  We use $\gamma=0.2$ for our simulations, which we found empirically to provide a good balance between equalization and reconstruction.

\section{Example Selection Algorithms}

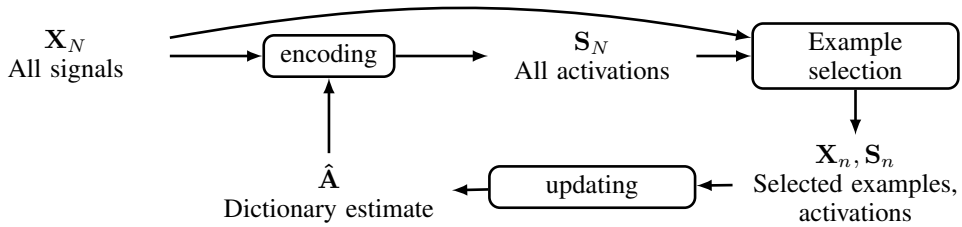
\begin{figure*}[t]
\centering
\begin{tikzpicture}[%
	every node/.style={align=center, text width=2.5cm, node distance=3.5cm and 0.5cm},
	every path/.style={-latex, line width=1pt}
]
	\node (XN) {${\bf X}_N$\\All signals};
	\node[right of=XN,draw,rounded corners,text width=1.5cm] (encoding) {encoding};
	\node[right of=encoding] (SN) {${\bf S}_N$\\All activations};
	\node[right of=SN,draw,rounded corners] (selection) {Example\\selection};
	\node[below=0.6cm of selection,text width=3.0cm] (Sn) {${\bf X}_n, {\bf S}_n$\\Selected examples,\\activations};
	\node[below=1cm of SN,draw,rounded corners] (learning) {updating};
	\node[below=1cm of encoding,text width=3.0cm] (Ahat) {${\bf {\hat A}}$\\Dictionary estimate};

	\draw (XN) -- (encoding);
	\draw (encoding) -- (SN);
	\draw (SN) -- (selection);
	\draw (selection) -- (Sn);
	\draw (Sn) -- (learning);
	\draw (learning) -- (Ahat);
	\draw (Ahat) -- (encoding);
	\draw[bend right] (XN) to[out=10,in=170] (selection);

\end{tikzpicture}
\caption{The interaction among encoding, selection and updating algorithms.}
\label{fig:flow_diagram}
\end{figure*}

To examine the effect of the example selection process on the learning, we extend the alternate optimization scheme in equations (\ref{eqn:encoding},~\ref{eqn:updating}) to include an \emph{example selection} stage.  In this stage, a selection algorithm picks $n \ll N$ examples to use for the dictionary update (Figure \ref{fig:flow_diagram}).  Ideally, the examples are to be chosen in such a way as to make learned dictionary ${\bf {\hat A}}$ closer to the ground-truth ${\bf A}^*$ compared to the uniform sampling. In the following, we describe a number of heuristic selection algorithms that were inspired by models of attention.

We characterize example selection algorithms in two parts.  First, there is a choice of \emph{goodness measure} $g_j$, which is a function that maps $({\bf s}^{(i)}, {\bf x}^{(i)})$ to a  number reflecting the ``goodness'' of the instance $i$ for the dictionary element $j$. Applying $g_j$ to $\{{\bf s}^{(i)}\}_{i=1}^N$ yields goodness values ${\bf G}_N$ for all $k$ dictionary elements and all $N$ examples. Second, there is a choice of \emph{selector} function $f_{sel}$.  This function dictates the way a subset of ${\bf X}_N$ is chosen using ${\bf G}_N$ values.

\subsection{Goodness measures}

Of the various goodness measures, we first consider

\begin{align}
 \textrm{\tt Err}: \quad g_j({\bf s}^{(i)}, {\bf x}^{(i)}) = \| {\bf {\hat A}}{\bf s}^{(i)} - {\bf x}^{(i)} \|_1 .
\end{align}

{\tt Err} is motivated by the idea of ``critical examples'' in \cite{Zhang:1994uy}, and it favors examples with large reconstruction errors.  In our paradigm, the criticality measured by {\tt Err} may not correspond to ground-truth errors, since it is calculated using current estimate ${\bf {\hat A}}$ rather than ground-truth ${\bf A}^*$.

Another related idea is to select examples that would produce large gradients in the dictionary update equation \eqref{eq:sgd}, without regard to their directions.  This results in

\begin{align}
 \textrm{\tt Grad}: \quad g_j({\bf s}^{(i)}, {\bf x}^{(i)}) = \| {\bf {\hat A}}{\bf s}^{(i)} - {\bf x}^{(i)} \|_1 \cdot s^{(i)}_j.
\end{align}

We note that {\tt Grad} extends {\tt Err} by multiplying the reconstruction errors by the activations $s^{(i)}_j$. It therefore prefers examples that are both critical and produce large activations.

One observation is that the level of noise puts a fundamental limit on the recovery of true dictionary: better approximation bound is obtained when observation noise is low. It follows that, if we can somehow collect examples that happen to have low noise, learning from those examples might be beneficial.  This motivated us to consider

\begin{align}
 \textrm{\tt SNR}: \quad g_j({\bf s}^{(i)}, {\bf x}^{(i)}) = \frac{\| {\bf x}^{(i)}\|_2^2}{\| {\bf {\hat A}}{\bf s}^{(i)} - {\bf x}^{(i)} \|_2^2} \cdot s^{(i)}_j.
\end{align}

This measure prefers examples with large estimated signal-to-noise ratio (SNR).

Another idea focuses on the statistical property of activations ${\bf s}^{(i)}$, inspired by a model of visual saliency proposed by \citet{Zhang2008}.  Their saliency model, called the SUN model, asserts that signals that result in rare feature activations are more salient. Specifically, the model defines the saliency of a particular visual location to be proportional the self-information of the feature activation, $-\log P(F = f)$.  Because we assume nonzero activations are exponentially distributed, this corresponds to

\begin{align}
 \textrm{\tt SUN}: \quad g_j({\bf s}^{(i)}, {\bf x}^{(i)}) = s^{(i)}_j \quad \left (\propto -\log P(s^{(i)}_j) \right ) .
\end{align}

We note that this model is not only simple, but also does not depend on ${\bf x}^{(i)}$ directly.  This makes {\tt SUN} attractive as a neurally implementable goodness measure.

Another saliency-based goodness measure is inspired by the visual saliency map model of \citet{Itti:2002tq}: 

\begin{align}
 \textrm{\tt SalMap}: \quad g_j({\bf s}^{(i)}, {\bf x}^{(i)}) = SaliencyMap({\bf x}^{(i)}).
\end{align}

In contrast to the {\tt SUN} measure, {\tt SalMap} depends only on ${\bf x}^{(i)}$. Consequently, {\tt SalMap} is impervious to changes in ${\bf {\hat A}}$.  Since the signals in our simulations are small monochrome patches, the ``saliency map'' we use only has a single-scale intensity channel and an orientation channel with four directions.

\subsection{Selector functions}

We consider two selector functions.  The first function chooses top $n$ examples with high goodness values across dictionary elements:

\begin{align}
 \textrm{\tt BySum}: \quad f_{sel}({\bf G}_N) = \textrm{top $n$ elements of}  \sum_{j=1}^K {\bf G}_j^{(i)}.
\end{align}

The second selector function,selects examples that are separately ``good'' for each dictionary element:  

\begin{align}
\begin{split}
 \textrm{\tt ByElement}: \quad f_{sel}({\bf G}_N) = \\ \{ \textrm{top $n / K$ elements of } {\bf G}_j^{(i)} \;|\; j \in 1 ... K  \} .
\end{split}
\end{align}

This is done by first sorting ${\bf G}_j^{(i)}$ for each $j$ and then picking top examples in a round-robin fashion, until $N$ examples are selected. Barring duplicates, this yields a set consisting of top $n / k$ elements of ${\bf G}_j^{(i)}$ for each element $j$.   Algorithm~\ref{alg:main} describes how these operations take place within each learning epoch.

In our simulations, we consider all possible combinations of the goodness measures and selector functions for the example selection algorithm, except for {\tt Err} and {\tt SalMap}. Since these two goodness measures do not produce different values for different dictionary element activations $s_j^{(i)}$, {\tt BySum} and {\tt ByElement} functions select equivalent example sets.

\begin{center}
\begin{algorithm}[b!]
\caption{Learning with example selection}\label{alg:main}
Initialize random ${\bf {\hat A}}_0 \in \mathcal{C}_{\bf A}$ from training examples\\
For $t=1$ to max. epochs:
\begin{enumerate}[noitemsep,topsep=0pt,parsep=0pt,partopsep=0pt]
	\item Obtain training set ${\bf X}_N = \{ {\bf x}^{(i)} \}_{i=1}^N$
    \item Encode ${\bf X}_N$: ${\bf S}_N = \{ f_{enc}({\bf x}^{(i)}; {\bf {\hat A}}) \}_{i=1}^N$
	\item Select $n$ ``good'' examples
\begin{itemize}[noitemsep,topsep=0pt,parsep=0pt,partopsep=0pt,label={--}]
	\item Calculate ${\bf G}_N = \{[g_j({\bf s}^{(i)}, {\bf x}^{(i)})]_{j=1 ... k} \}_{i=1}^N$
	\item Select $n$ indices: $\Gamma = f_{sel}( {\bf G}_N )$
	\item  ${\bf S}_n = \{{\bf s}^{(i)}\}_{i \in \Gamma}$, ${\bf X}_n = \{{\bf x}^{(i)}\}_{i \in \Gamma}$

\end{itemize}
	\item Loop 10 times:
\begin{enumerate}[noitemsep,topsep=0pt,parsep=0pt,partopsep=0pt]
	\item Encode ${\bf X}_n$: ${\bf S}_n \leftarrow \{ f_{enc}({\bf x}^{(i)}; {\bf {\hat A}}) \}_{i=1}^n$
	\item Equalize ${\bf S}_n$: $\forall {\bf s}^{(i)} \in {\bf S}_n$, \\$s^{(i)}_j \leftarrow s^{(i)}_j \cdot (\frac{1}{K} \sum_{j=1}^K \sum_{i=1}^n s^{(i)}_j  / \sum_{i=1}^n s^{(i)}_j )^{\gamma}$ 
	\item Update ${\bf {\hat A}}$: ${\bf {\hat A}} \leftarrow {\bf {\hat A}} - \eta_t ({\bf {\hat A}} {\bf S}_n-{\bf X}_n) {\bf S}_n^\intercal / n$
	\item Normalize columns of ${\bf {\hat A}}$.
\end{enumerate}
\end{enumerate}
\end{algorithm}
\end{center}

\section{Simulations}

In order to evaluate example selection algorithms, we present simulations across a variety of dictionaries and encoding algorithms. Specifically, we compare results using all three possible encoding models ({\tt L0}, {\tt L1}, and {\tt k-Sparse}) with all eight selection algorithms.  Because we generate the training examples from a known ground-truth dictionary ${\bf A}^*$, we quantify the integrity of learned dictionary ${\bf {\hat A}}_t$ at each learning epoch $t$ using the minimal mean square distance

\begin{align}
D^*({\bf {\hat A}}, {\bf A}^*) \triangleq \min_{{\bf P}_\pi} \frac{1}{KP} \| {\bf {\hat A}}_t {\bf P}_\pi  - {\bf A}^* \|_F^2 ,
\end{align}

\noindent with ${\bf P}_\pi$ spanning all possible permutations.

\begin{figure*}[t]
  \begin{subfigure}[b]{0.5\textwidth}
    \centering
    \includegraphics[width=0.4\textwidth]{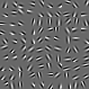}
    \includegraphics[width=0.4\textwidth]{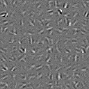}
    \caption{Gabor dictionary}
    \label{fig:dic_gabors}
  \end{subfigure}%
  ~ 
  \begin{subfigure}[b]{0.5\textwidth}
    \centering
    \includegraphics[width=0.4\textwidth]{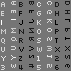}
    \includegraphics[width=0.4\textwidth]{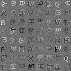}
    \caption{Alphanumeric dictionary}
    \label{fig:dic_letters}
  \end{subfigure}
  \caption{Ground-truth dictionaries and generated examples ${\bf X}_N$. Each element / generated example is a 8x8 patch, displayed as a tiled image for the ease of visualization. White is positive and black is negative.}\label{fig:dictionaries}
\end{figure*}

We also investigate the effect of ${\bf A}^*$ on the learning.  One way to characterize a dictionary set ${\bf A}$ is its mutual coherence $\mu({\bf A}) \triangleq \max_{i \neq j} | {\bf a}_i^\intercal {\bf a}_j|$~\cite{elad}.  This measure is useful in theoretical analysis of recovery bounds~\cite{donoho}. A more practical characterization is the average coherence  ${\bar \mu}({\bf A}) \triangleq \frac{2}{K(K-1)} \sum_{i \neq j} |{\bf a}_i^\intercal {\bf a}_j|$. Regardless, exact recovery of the dictionary is more challenging when the coherence is high.

The first dictionary set comprises $100$ $8$x$8$ Gabor patches (Figure~ \ref{fig:dic_gabors}).  This dictionary set is inspired by the fact that dictionary learning of natural images leads to such a dictionary~\cite{Olshausen:1996p2797}, and they correspond to simple receptive fields in mammalian visual cortices~\cite{Jones:1987ub}. With $\mu({\bf A}^*) = 0.97$ but ${\bar \mu}({\bf A}^*) = 0.13$, this dictionary set is relatively incoherent, and so the learning problem should be easier.

The second dictionary set is composed of $64$ $8$x$8$ alphanumeric letters with alternating rotations and signs (Figure~\ref{fig:dic_letters}).  This artificial dictionary set has $\mu({\bf A}^*) = 0.95$ with ${\bar \mu}({\bf A}^*) = 0.34$\footnote{Both dictionaries violate the recovery bound described in \cite{donoho}.  \citet{Amiri:2014ct} notes that this bound is prone to be violated in practice; as such, we explicitly chose ``realistic'' parameters that violate the bounds in our simulations.}.

Within each epoch, 50,000 examples are generated with 5 nonzero activations per example ($k=5$), whose magnitudes are sampled from $Exp(1)$. $\sigma^2_\epsilon$ is set so that examples have SNR of $\approx 6$ dB. Each selection algorithm then picks 1\% ($n=500$) of the training set for the learning. For each experiment, ${\bf {\hat A}}$ is initialized with random examples from the training set.


\subsection{Results}

Figure~\ref{fig:res_dist} shows the average distance of ${\bf {\hat A}}$ from ${\bf A}^*$ for each learning epoch. We observe that {\tt ByElement} selection policies generally work well, especially in conjunction with {\tt Grad} and {\tt SUN} goodness measures.  This trend is especially noticeable for the alphanumeric dictionary case, where most of the {\tt BySum}-selectors perform worse than the baseline selector that chooses examples randomly ({\tt Uniform}).

The ranking of the selector algorithms is roughly consistent across the learning epochs (Figure~\ref{fig:res_dist}, left column), and it is also robust with the choice of the encoding algorithms  (Figure~\ref{fig:res_dist}, right column). In particular, good selector algorithms are beneficial even at the relatively early stages of learning ($< 100$ epochs, for instance), in contrast to the simulation in \cite{Amiri:2014ct}.  This is surprising, because at early stages of learning, poor ${\bf {\hat A}}$ estimates result in bad activation estimates as well. Nevertheless, good selector algorithms soon establish a positive feedback loop for both dictionary and activation estimates.

One interesting exception is the {\tt SalMap} selector. It works relatively well for Gabor dictionary (and closely tracks the {\tt SUNBySum} selector), but not for the alphanumeric dictionary.  This is presumably due to the design of the {\tt SalMap} model: because the model uses oriented Gabor filters as one of its feature maps, the overall effect is similar to the {\tt SUNBySum} algorithm when the signals are generated from Gabor dictionaries.

\setlength\figureheight{0.35\textwidth}

\begin{figure*}[ht]
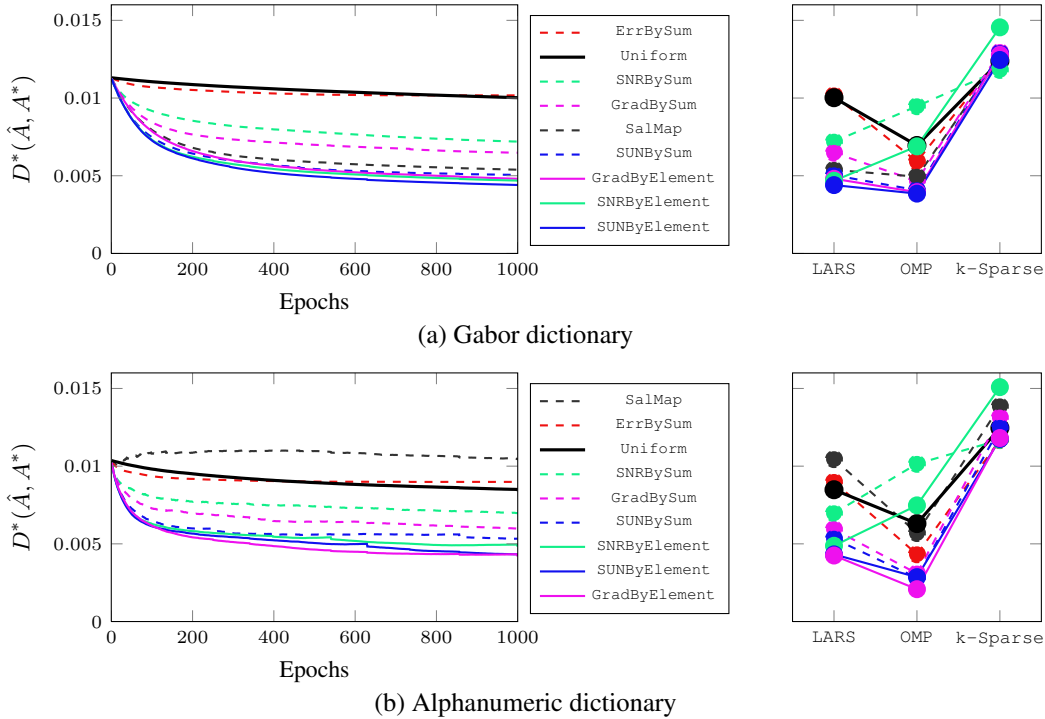

\begin{tabular}{rl}
\setlength\figurewidth{0.5\textwidth}
\imagetop{\input{nips-gabors--dist_A-only_lasso-for_paper.tikz}} &
\setlength\figurewidth{0.35\textwidth}
\imagetop{
%
%
%
%
\begin{tikzpicture}

\definecolor{color1}{rgb}{0.0666666666666667,0.933333333333333,0.533333333333333}
\definecolor{color0}{rgb}{0.933333333333333,0.0666666666666667,0.0666666666666667}
\definecolor{color3}{rgb}{0.0666666666666667,0.0666666666666667,0.933333333333333}
\definecolor{color2}{rgb}{0.933333333333333,0.0666666666666667,0.933333333333333}

\begin{axis}[
xmin=-0.5, xmax=2.5,
ymin=0, ymax=0.016,
axis on top,
width=\figurewidth,
height=\figureheight,
xtick={0,1,2},
xticklabels={\tt LARS,\tt OMP,\tt k-Sparse},
yticklabels={},
x tick label style={/pgf/number format/.cd, precision=3, fixed, 1000 sep={}},
y tick label style={/pgf/number format/.cd, precision=3, fixed, 1000 sep={}},
scaled y ticks=false,
scaled x ticks=false
]
\addplot [thick, color0, dashed, mark=*, mark size=3]
coordinates {
(-1.11022302462516e-16,0.0101712058324572)
(1,0.00590844868237794)
(2,0.0126194837854742)

};
\addplot [very thick, black, mark=*, mark size=3]
coordinates {
(-1.11022302462516e-16,0.0100225295718439)
(1,0.00694749486225738)
(2,0.0123774938106847)

};
\addplot [thick, color1, dashed, mark=*, mark size=3]
coordinates {
(-1.11022302462516e-16,0.00719487343976466)
(1,0.00947322712291063)
(2,0.011810174428342)

};
\addplot [thick, color2, dashed, mark=*, mark size=3]
coordinates {
(-1.11022302462516e-16,0.00647652056523248)
(1,0.00466044849787685)
(2,0.012960652627834)

};
\addplot [thick, white!20.0!black, dashed, mark=*, mark size=3]
coordinates {
(-1.11022302462516e-16,0.00537829864286787)
(1,0.00494012761642477)
(2,0.0125771047151826)

};
\addplot [thick, color3, dashed, mark=*, mark size=3]
coordinates {
(-1.11022302462516e-16,0.00503236777474787)
(1,0.0040698086993132)
(2,0.0128914139318871)

};
\addplot [thick, color2, mark=*, mark size=3]
coordinates {
(-1.11022302462516e-16,0.00480449076012392)
(1,0.00394395691770288)
(2,0.0127799886300557)

};
\addplot [thick, color1, mark=*, mark size=3]
coordinates {
(-1.11022302462516e-16,0.00467742579458158)
(1,0.00689153818348837)
(2,0.0145498942574049)

};
\addplot [thick, color3, mark=*, mark size=3]
coordinates {
(-1.11022302462516e-16,0.00439916169779191)
(1,0.00385022015369705)
(2,0.0124515503863632)

};

\end{axis}

\end{tikzpicture}} \\
\multicolumn{2}{c}{(a) Gabor dictionary} \\
\setlength\figurewidth{0.5\textwidth}
\imagetop{\input{nips-letters--dist_A-only_lasso-for_paper.tikz}} &
\setlength\figurewidth{0.35\textwidth}
\imagetop{
%
%
%
%
\begin{tikzpicture}

\definecolor{color1}{rgb}{0.0666666666666667,0.933333333333333,0.533333333333333}
\definecolor{color0}{rgb}{0.933333333333333,0.0666666666666667,0.0666666666666667}
\definecolor{color3}{rgb}{0.0666666666666667,0.0666666666666667,0.933333333333333}
\definecolor{color2}{rgb}{0.933333333333333,0.0666666666666667,0.933333333333333}

\begin{axis}[
xmin=-0.5, xmax=2.5,
ymin=0, ymax=0.016,
axis on top,
width=\figurewidth,
height=\figureheight,
xtick={0,1,2},
xticklabels={\tt LARS,\tt OMP,\tt k-Sparse},
yticklabels={},
x tick label style={/pgf/number format/.cd, precision=3, fixed, 1000 sep={}},
y tick label style={/pgf/number format/.cd, precision=3, fixed, 1000 sep={}},
scaled y ticks=false,
scaled x ticks=false
]
\addplot [thick, white!20.0!black, dashed, mark=*, mark size=3]
coordinates {
(-1.11022302462516e-16,0.0104516377130232)
(1,0.00569474164697562)
(2,0.0138497799500303)

};
\addplot [thick, color0, dashed, mark=*, mark size=3]
coordinates {
(-1.11022302462516e-16,0.00898711542116323)
(1,0.00432915763502883)
(2,0.0116748889112658)

};
\addplot [very thick, black, mark=*, mark size=3]
coordinates {
(-1.11022302462516e-16,0.00848295506976741)
(1,0.00630452151302348)
(2,0.0124602825509883)

};
\addplot [thick, color1, dashed, mark=*, mark size=3]
coordinates {
(-1.11022302462516e-16,0.0069703934276349)
(1,0.0101493860678341)
(2,0.0116884351159178)

};
\addplot [thick, color2, dashed, mark=*, mark size=3]
coordinates {
(-1.11022302462516e-16,0.00593815765787194)
(1,0.00308979012700844)
(2,0.0131283215517417)

};
\addplot [thick, color3, dashed, mark=*, mark size=3]
coordinates {
(-1.11022302462516e-16,0.00533385325277604)
(1,0.00281589524501494)
(2,0.0124663834148758)

};
\addplot [thick, color1, mark=*, mark size=3]
coordinates {
(-1.11022302462516e-16,0.0048758119865242)
(1,0.00747651362646244)
(2,0.0150850221423615)

};
\addplot [thick, color3, mark=*, mark size=3]
coordinates {
(-1.11022302462516e-16,0.00432276990650476)
(1,0.00285728164238401)
(2,0.0117495631403755)

};
\addplot [thick, color2, mark=*, mark size=3]
coordinates {
(-1.11022302462516e-16,0.00424902556621943)
(1,0.00208199940644048)
(2,0.0118113219596468)

};

\end{axis}

\end{tikzpicture}} \\
\multicolumn{2}{c}{(b) Alphanumeric dictionary} \\
\end{tabular}
\caption{Distance from true dictionaries. Graphs on the left column show the time course of the learning using the {\tt LARS} encoding.  The legends are ordered from worst to best at the end of the simulation (1000 epochs). Graphs on the right column compares the performance of different encoding models.  The ordinate is the distance at the end, in the same scale as the left graphs.}
\label{fig:res_dist}
\end{figure*}

\subsection{Robustness}

In order to assess the robustness of the example selection algorithms, we repeated the Gabor dictionary simulation across a range of parameter values. Specifically, we experimented with modifying the following parameters one at a time, starting from the original parameter values:

\begin{itemize}
	\item The signal-to-noise ratio ($10 \log_{10} (2 \lambda^2 / \sigma_\epsilon^2)$ [dB])
	\item The number of nonzero elements in the generated examples ($k$)
	\item The ratio of selected examples to the original training set ($n / N$)
	\item The number of dictionary elements ($K$)
\end{itemize}

Figure~\ref{fig:params} shows the result of these simulations. These results show that good selector algorithms improve learning across a wide range of parameter values.  Of note is  the number of dictionary elements $K$, whose results suggest that the improvement is greatest for the ``complete'' dictionary learning cases; the advantage of selection appears to diminish for extremely over-complete (or under-complete) dictionary learning tasks.

\setlength\figureheight{0.25\textwidth}
\setlength\figurewidth{0.35\textwidth}

\begin{figure*}[ht]
\centering
\begin{tabular}{rl}
\imagetop{
%
%
%
%
\begin{tikzpicture}

\definecolor{color1}{rgb}{0.0666666666666667,0.933333333333333,0.533333333333333}
\definecolor{color0}{rgb}{0.933333333333333,0.0666666666666667,0.0666666666666667}
\definecolor{color3}{rgb}{0.0666666666666667,0.0666666666666667,0.933333333333333}
\definecolor{color2}{rgb}{0.933333333333333,0.0666666666666667,0.933333333333333}

\begin{axis}[
xlabel={SNR [dB]},
ylabel={$D^*(\boldsymbol{\hat A}, \boldsymbol{A}^*)$},
xmin=-18, xmax=-12,
ymin=0, ymax=0.016,
axis on top,
width=\figurewidth,
height=\figureheight,
x tick label style={/pgf/number format/.cd, precision=3, fixed, 1000 sep={}},
y tick label style={/pgf/number format/.cd, precision=3, fixed, 1000 sep={}},
scaled y ticks=false,
scaled x ticks=false
]
\addplot [thick, color0, dashed]
coordinates {
(-18,0.0118262035084913)
(-16.5,0.0104173239924024)
(-15,0.0100634790538966)
(-13.5,0.0102731831746069)
(-12,0.0109278456209119)

};
\addplot [very thick, black]
coordinates {
(-18,0.0109050284663039)
(-16.5,0.0102779184792723)
(-15,0.0103177191603829)
(-13.5,0.010489941648549)
(-12,0.0108357557725376)

};
\addplot [thick, color1, dashed]
coordinates {
(-18,0.00527893049513762)
(-16.5,0.0061228020388247)
(-15,0.00795407667761272)
(-13.5,0.00788069314728451)
(-12,0.00794540226728969)

};
\addplot [thick, color2, dashed]
coordinates {
(-18,0.00600137204864951)
(-16.5,0.0052620153969324)
(-15,0.00611227101221857)
(-13.5,0.00636178676382026)
(-12,0.00711951303485407)

};
\addplot [thick, white!20.0!black, dashed]
coordinates {
(-18,0.00738588189839336)
(-16.5,0.0059421466118604)
(-15,0.00602228959420366)
(-13.5,0.00548792292301796)
(-12,0.00596131060207544)

};
\addplot [thick, color2]
coordinates {
(-18,0.00588770944240371)
(-16.5,0.00405781306979138)
(-15,0.00496051660159298)
(-13.5,0.00472320973176314)
(-12,0.00484424307500645)

};
\addplot [thick, color3, dashed]
coordinates {
(-18,0.00450327128799242)
(-16.5,0.00439051852769474)
(-15,0.00466501896721145)
(-13.5,0.00529360036040412)
(-12,0.00516284462489772)

};
\addplot [thick, color1]
coordinates {
(-18,0.00413394542013683)
(-16.5,0.00398130023054811)
(-15,0.00454506611078481)
(-13.5,0.00420629675951404)
(-12,0.00517848478914456)

};
\addplot [thick, color3]
coordinates {
(-18,0.00412332439079001)
(-16.5,0.00400588430659155)
(-15,0.00453757326461833)
(-13.5,0.00422109510615427)
(-12,0.00505839785548912)

};

\end{axis}

\end{tikzpicture}} &
\imagetop{
%
%
%
%
\begin{tikzpicture}

\definecolor{color1}{rgb}{0.0666666666666667,0.933333333333333,0.533333333333333}
\definecolor{color0}{rgb}{0.933333333333333,0.0666666666666667,0.0666666666666667}
\definecolor{color3}{rgb}{0.0666666666666667,0.0666666666666667,0.933333333333333}
\definecolor{color2}{rgb}{0.933333333333333,0.0666666666666667,0.933333333333333}

\begin{axis}[
xlabel={$k$},
ylabel={$D^*(\boldsymbol{\hat A}, \boldsymbol{A}^*)$},
xmin=2, xmax=10,
ymin=0, ymax=0.016,
axis on top,
width=\figurewidth,
height=\figureheight,
x tick label style={/pgf/number format/.cd, precision=3, fixed, 1000 sep={}},
y tick label style={/pgf/number format/.cd, precision=3, fixed, 1000 sep={}},
legend entries={{\tt ErrBySum},{\tt Uniform},{\tt SNRBySum},{\tt GradBySum},{\tt SalMap},{\tt GradByElement},{\tt SUNBySum},{\tt SNRByElement},{\tt SUNByElement}},
scaled y ticks=false,
scaled x ticks=false,
legend style={at={(1.13,1.03)}, anchor=north west}
]
\addplot [thick, color0, dashed]
coordinates {
(2,0.00989207159028167)
(4,0.010274381991671)
(6,0.010990036694002)
(8,0.0106549851401582)
(10,0.00979200903881197)

};
\addplot [very thick, black]
coordinates {
(2,0.010979822265243)
(4,0.00991157052278847)
(6,0.010699587960254)
(8,0.00997760335725307)
(10,0.00909469396921722)

};
\addplot [thick, color1, dashed]
coordinates {
(2,0.00844179315840295)
(4,0.00769355767142722)
(6,0.00791988154212403)
(8,0.00798015863495254)
(10,0.00763659512201445)

};
\addplot [thick, color2, dashed]
coordinates {
(2,0.00712673057015407)
(4,0.00696143323849892)
(6,0.00688105666642567)
(8,0.00672951758769454)
(10,0.0060809315438471)

};
\addplot [thick, white!20.0!black, dashed]
coordinates {
(2,0.00615937962551268)
(4,0.00540769049582938)
(6,0.00641983352730397)
(8,0.00621128344855361)
(10,0.00500910192108511)

};
\addplot [thick, color2]
coordinates {
(2,0.00576394289259989)
(4,0.00536104012032502)
(6,0.00500574911589747)
(8,0.00561096555484936)
(10,0.00462863384035637)

};
\addplot [thick, color3, dashed]
coordinates {
(2,0.00599768411374703)
(4,0.00514383387008247)
(6,0.00565614276657125)
(8,0.00524423301404086)
(10,0.00410188122954347)

};
\addplot [thick, color1]
coordinates {
(2,0.00575617607579973)
(4,0.00475471007169303)
(6,0.00500710674203368)
(8,0.00493397738374598)
(10,0.00374677512268749)

};
\addplot [thick, color3]
coordinates {
(2,0.00592207049386681)
(4,0.00470020402291718)
(6,0.00495022558850894)
(8,0.00469538348999714)
(10,0.00371606265178283)

};

\path[draw=black, dotted] (axis cs:5,0)--(axis cs:5,0.016);

\end{axis}

\end{tikzpicture}} \\
\imagetop{
%
%
%
%
\begin{tikzpicture}

\definecolor{color1}{rgb}{0.0666666666666667,0.933333333333333,0.533333333333333}
\definecolor{color0}{rgb}{0.933333333333333,0.0666666666666667,0.0666666666666667}
\definecolor{color3}{rgb}{0.0666666666666667,0.0666666666666667,0.933333333333333}
\definecolor{color2}{rgb}{0.933333333333333,0.0666666666666667,0.933333333333333}

\begin{semilogxaxis}[
xlabel={$n/N$},
ylabel={$D^*(\boldsymbol{\hat A}, \boldsymbol{A}^*)$},
xmin=0.01, xmax=1,
ymin=0, ymax=0.016,
axis on top,
width=\figurewidth,
height=\figureheight,
x tick label style={/pgf/number format/.cd, precision=2, fixed},
xticklabels={$0.01$, $0.01$, $0.1$, $1$},
y tick label style={/pgf/number format/.cd, precision=3, fixed, 1000 sep={}},
scaled y ticks=false,
scaled x ticks=false,
]
\addplot [very thick, black]
coordinates {
(0.01,0.0107532205252918)
(0.0298538261891796,0.0105046056953761)
(0.0891250938133746,0.0107357207288275)
(0.266072505979881,0.0106711226945071)
(1, 0.0107)
};
\addplot [thick, color0, dashed]
coordinates {
(0.01,0.0102184265144015)
(0.0298538261891796,0.0102033502088081)
(0.0891250938133746,0.0101923707915327)
(0.266072505979881,0.0102315702942135)
(1, 0.0107)

};
\addplot [thick, color1, dashed]
coordinates {
(0.01,0.00755640660198595)
(0.0298538261891796,0.00748064505724256)
(0.0891250938133746,0.00688100060008747)
(0.266072505979881,0.00885097317879067)
(1, 0.0107)

};
\addplot [thick, color2, dashed]
coordinates {
(0.01,0.00698272122228204)
(0.0298538261891796,0.00674314639844991)
(0.0891250938133746,0.00695539169101109)
(0.266072505979881,0.00884275134775631)
(1, 0.0107)

};
\addplot [thick, white!20.0!black, dashed]
coordinates {
(0.01,0.00685508058255469)
(0.0298538261891796,0.00619836855694646)
(0.0891250938133746,0.00702759444020219)
(0.266072505979881,0.00891503892722001)
(1, 0.0107)

};
\addplot [thick, color2]
coordinates {
(0.01,0.00562334953778681)
(0.0298538261891796,0.00611421403491156)
(0.0891250938133746,0.00682406091686668)
(0.266072505979881,0.00884280078999129)
(1, 0.0107)

};
\addplot [thick, color3, dashed]
coordinates {
(0.01,0.00501290745515904)
(0.0298538261891796,0.00613519263915294)
(0.0891250938133746,0.00669716858661465)
(0.266072505979881,0.00884278608144329)
(1, 0.0107)

};
\addplot [thick, color1]
coordinates {
(0.01,0.00511132040063875)
(0.0298538261891796,0.0058813109143959)
(0.0891250938133746,0.00668196681279637)
(0.266072505979881,0.00884274238403542)
(1, 0.0107)

};
\addplot [thick, color3]
coordinates {
(0.01,0.00516815360209475)
(0.0298538261891796,0.00583271709715281)
(0.0891250938133746,0.00667350766731611)
(0.266072505979881,0.00884279019478922)
(1, 0.0107)

};

\path [draw=red, dotted, fill opacity=0] (axis cs:0,0)--(axis cs:0,0.016);

\end{semilogxaxis}

\end{tikzpicture}} &
\imagetop{
%
%
%
%
\begin{tikzpicture}

\definecolor{color1}{rgb}{0.933333333333333,0.0666666666666667,0.0666666666666667}
\definecolor{color0}{rgb}{0.0666666666666667,0.933333333333333,0.533333333333333}
\definecolor{color3}{rgb}{0.0666666666666667,0.0666666666666667,0.933333333333333}
\definecolor{color2}{rgb}{0.933333333333333,0.0666666666666667,0.933333333333333}

\begin{semilogxaxis}[
xlabel={$K$},
ylabel={$D^*(\boldsymbol{\hat A}, \boldsymbol{A}^*)$},
xmin=8, xmax=256,
ymin=0, ymax=0.016,
axis on top,
width=\figurewidth,
height=\figureheight,
x tick label style={/pgf/number format/.cd, precision=3, fixed, 1000 sep={}},
y tick label style={/pgf/number format/.cd, precision=3, fixed, 1000 sep={}},
xtick = {8, 16, 32, 64, 128, 256},
xticklabels = {$8$, $16$, $32$, $64$, $128$, $256$},
scaled y ticks=false,
scaled x ticks=false
]
\addplot [thick, color0, dashed]
coordinates {
(8,0.0076459272673916)
(16,0.00968819865233274)
(32,0.00861408482230225)
(64,0.00894945545181599)
(128,0.00802797120399936)
(256,0.0089561108621007)

};
\addplot [thick, color1, dashed]
coordinates {
(8,0.0018183811207214)
(16,0.00718330542805881)
(32,0.00770814542295088)
(64,0.0111612435682012)
(128,0.0107627902913831)
(256,0.0107039254098877)

};
\addplot [very thick, black]
coordinates {
(8,0.00491654044407403)
(16,0.00572187481326862)
(32,0.00538372736042179)
(64,0.0105103906842207)
(128,0.0104807581106588)
(256,0.0108396681341525)

};
\addplot [thick, color2, dashed]
coordinates {
(8,0.00490373697631323)
(16,0.00612567949594099)
(32,0.00274170015005715)
(64,0.00645760669300355)
(128,0.00704865390353637)
(256,0.00864565230095441)

};
\addplot [thick, white!20.0!black, dashed]
coordinates {
(8,0.00504772563260182)
(16,0.00388634446359664)
(32,0.00281191373132144)
(64,0.00580053037270283)
(128,0.00596251469556158)
(256,0.00832037650937847)

};
\addplot [thick, color2]
coordinates {
(8,0.00540440603404029)
(16,0.00673271214359984)
(32,0.00161833031068389)
(64,0.00433968790701929)
(128,0.00511500506918538)
(256,0.00807342512376106)

};
\addplot [thick, color3, dashed]
coordinates {
(8,0.00495694236421196)
(16,0.00599758580957618)
(32,0.00304116833751549)
(64,0.00354912651358542)
(128,0.00541158658358367)
(256,0.00768209156435249)

};
\addplot [thick, color0]
coordinates {
(8,0.0051286782531426)
(16,0.00437145087118938)
(32,0.00347929191586348)
(64,0.00329584180300811)
(128,0.00495668941933391)
(256,0.00761478127321381)

};
\addplot [thick, color3]
coordinates {
(8,0.00506800323673483)
(16,0.00452532429438306)
(32,0.00236416666227222)
(64,0.00311745650908017)
(128,0.00486693447721705)
(256,0.00754202027056997)

};

\path[draw=black, dotted] (axis cs:100,0)--(axis cs:100,0.016);
\end{semilogxaxis}

\end{tikzpicture}} \\

\end{tabular}

  \caption{Distances from the true dictionaries for different model parameters, using the {\tt LARS} encoding.}
  \label{fig:params}
\end{figure*}
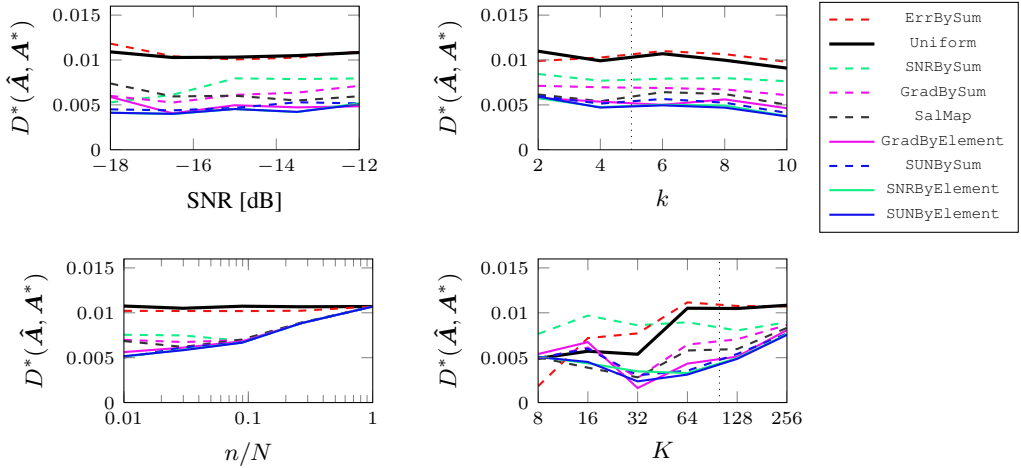

\section{Discussion}

In this work, we examined the effect of selection algorithms on the dictionary learning based on stochastic gradient descent.  Simulations using training examples generated from known dictionaries revealed that some selection algorithms do indeed improve learning, in the sense that the learned dictionaries are closer to the known dictionaries throughout the learning epochs. Of special note is the success of {\tt SUN} selectors; since these selectors are very simple, they hold promise for more general learning applications.

Few studies have so far investigated example selection strategies for the dictionary learning task, although some learning algorithms contain such procedures implicitly.  For instance, K-SVD \cite{Aharon:2006kn} relies upon identifying a group of examples that use a particular dictionary element during its update stage.  The algorithm in~\cite{Arora:2013vq} also makes use of a sophisticated example grouping procedure to provably recover dictionaries.  In both cases, though, the focus is on breaking the inter-dependency between ${\bf {\hat A}}$ and ${\bf {\hat S}}$, instead of characterizing how some algorithms -- notably those of the perceptual systems -- might improve learning despite this inter-dependency.

One recent paper that does consider example selection on its own is \citep{Amiri:2014ct}, whose {\tt cognit} algorithm is explicitly related to perceptual attention.  The point that differentiates this work lies in the generative assumption: {\tt cognit} relies on having additional information available to the learner, in their case the temporal contiguity of the generative process. With a spatially and temporally independent generation process, the generative model we considered here is simpler but more difficult to solve.

Why do selection algorithms improve learning at all?  At first glance, one may assume that any non-uniform sampling would skew the apparent distribution $\mathcal{D}({\bf X}_n)$ from the true distribution of the training set $\mathcal{D}({\bf X}_N)$, and thus lead to learning of an incorrect dictionary. However, as we have empirically shown, this is not the case. One intuitive reason -- one that also underlies the design of the {\tt SNR} selectors -- is that ``good'' selection algorithms picks samples with high information content.  For instance, samples with close to zero activation content provide little information about the dictionary elements that compose them, even though such samples abound under our generative model with exponentially-distributed activations. It follows that such samples provide little benefit to the inference of the statistical structure of the training set, and the learner would be well-advised to discard them.

To validate this, we calculated the (true) SNR of ${\bf X}_n$ at the last epoch of the learning for each selection algorithm (Figure~\ref{fig:snr}, left columns).  This shows that all selection algorithms picked ${\bf X}_n$ with much higher SNR than {\tt Uniform}.  However, the correlation between the overall performance ranking and SNR is weak, suggesting that this is not the only factor driving good example selection.

Another factor that contributes to good learning is the spread of examples within ${\bf X}_n$.  Casual observation revealed that the {\tt BySum} selector is prone to picking similar examples, whereas {\tt ByElement} selects a larger variety of examples and thus retains the distribution of ${\bf X}_N$ more faithfully.  To quantify this, we measured the distance of the distribution of selected examples, $\mathcal{D}({\bf X}_n)$, from that of all training examples, $\mathcal{D}({\bf X}_N)$, using the histogram intersection distance\cite{Rubner:2000uj}. The right columns of Figure~\ref{fig:snr} shows that this distance, $D(\mathcal{D}({\bf X}_n) || \mathcal{D}({\bf X}_N))$, tends to be lower for {\tt ByElement} selectors (solid lines) than {\tt BySum} selectors (dashed lines). Like the SNR measure, however, this quantity itself is only weakly predictive of the overall performance, suggesting that it is important to pick a large variety of high-SNR examples for the dictionary learning task.

\setlength\figureheight{0.28\textwidth}
\setlength\figurewidth{0.25\textwidth}
\begin{figure*}
\centering
\begin{tabular}{rlrl}
\imagetop{\includegraphics[width=\figurewidth,height=\figureheight,keepaspectratio]{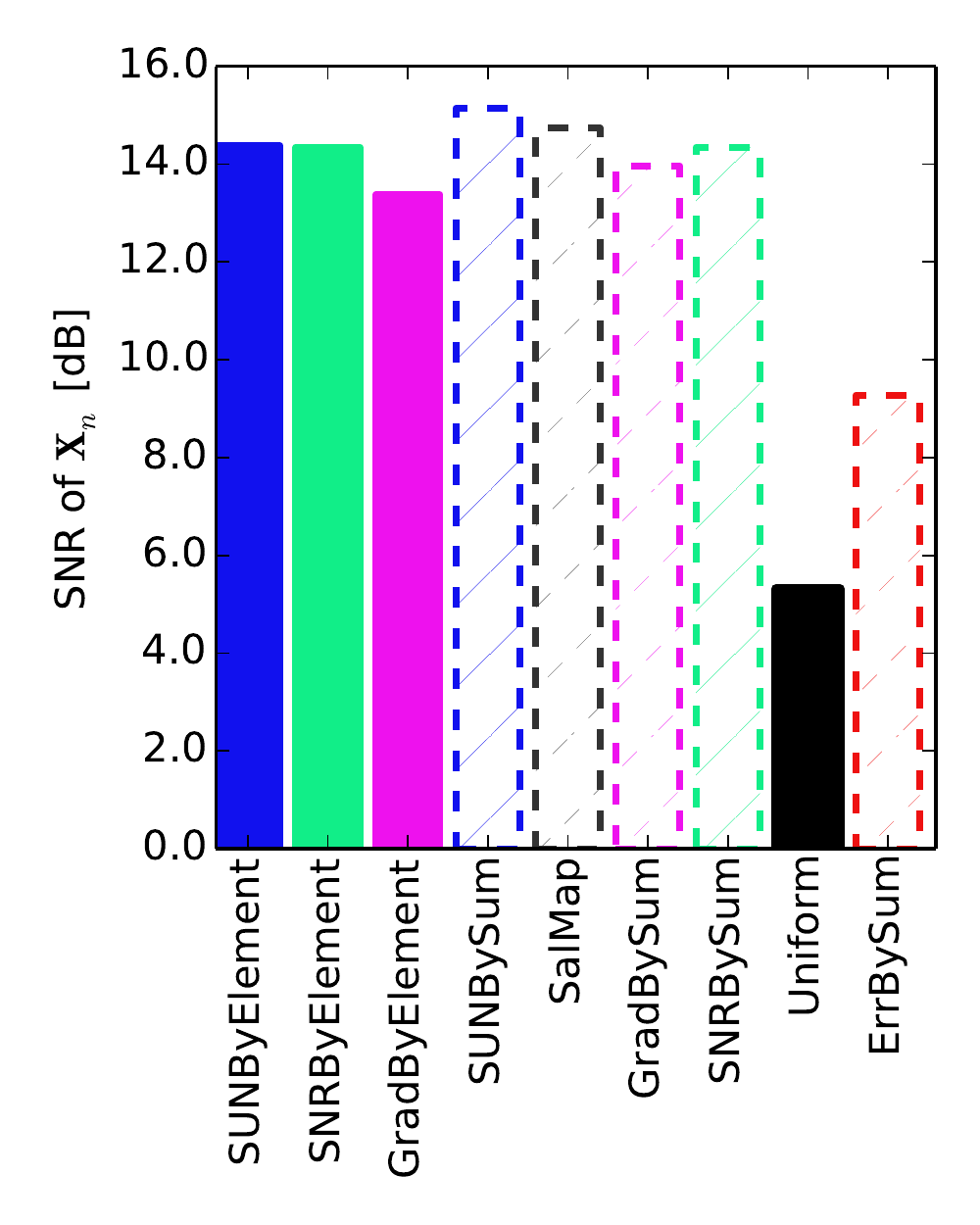}} &
\imagetop{\includegraphics[width=\figurewidth,height=\figureheight,keepaspectratio]{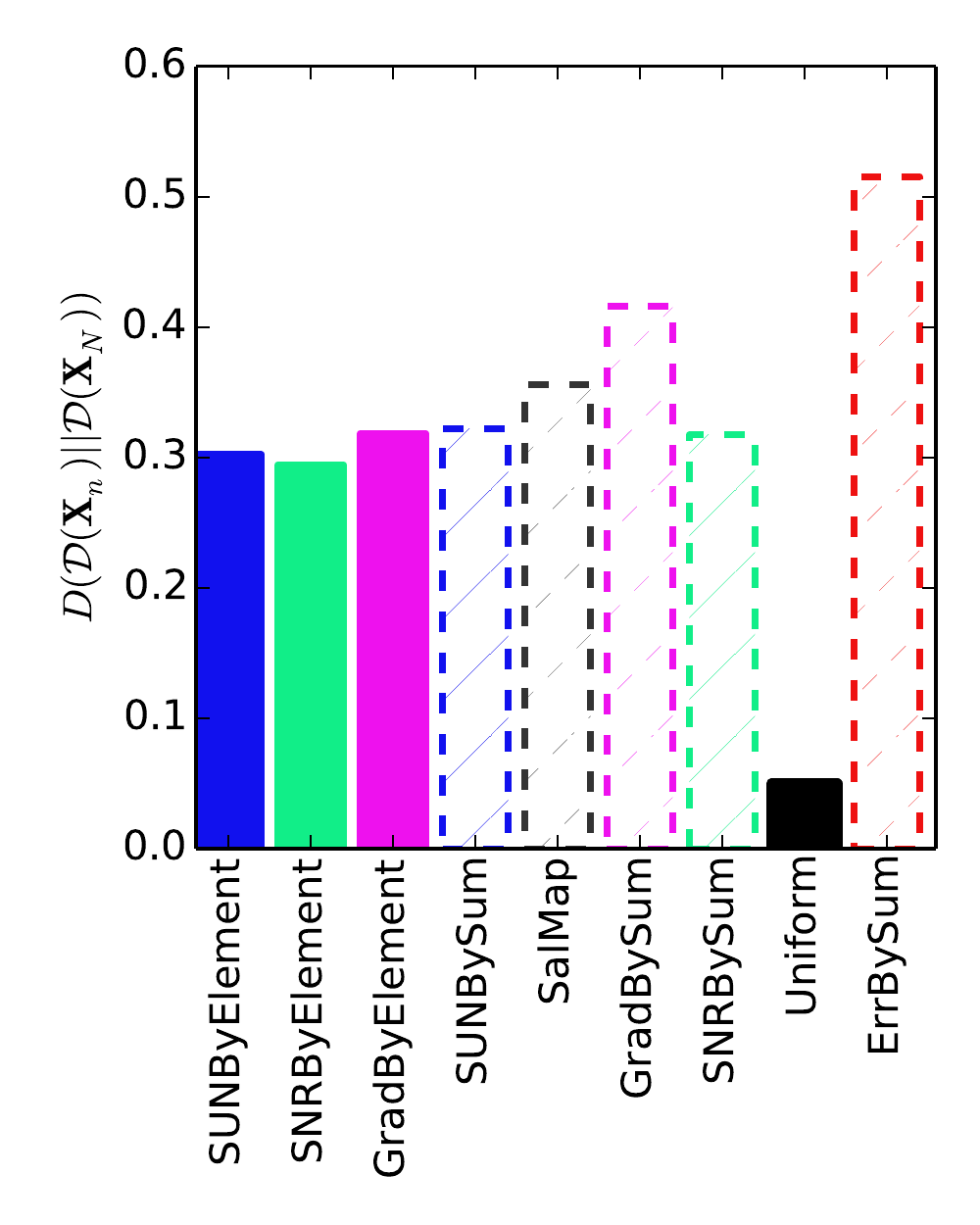}} &
\imagetop{\includegraphics[width=\figurewidth,height=\figureheight,keepaspectratio]{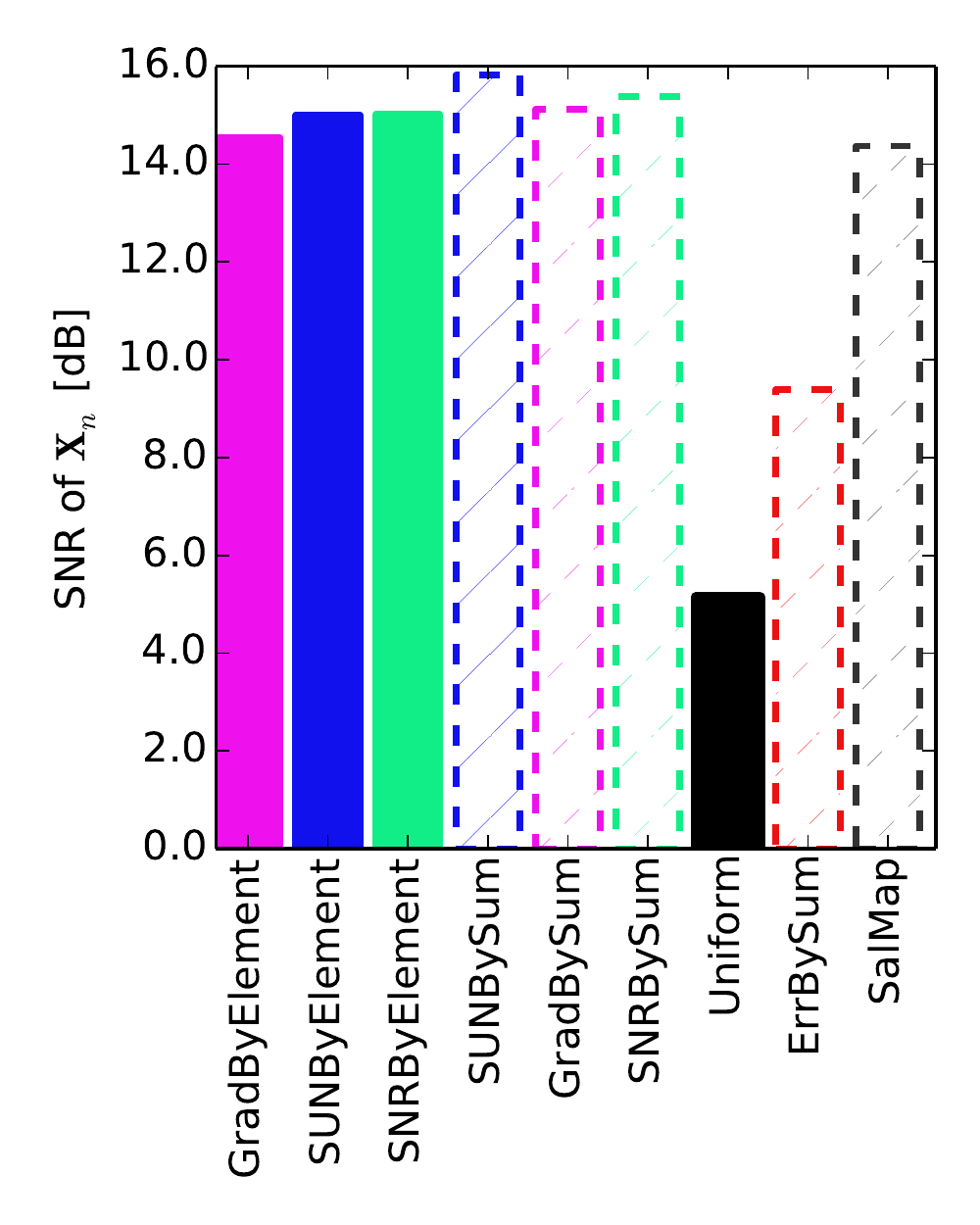}} &
\imagetop{\includegraphics[width=\figurewidth,height=\figureheight,keepaspectratio]{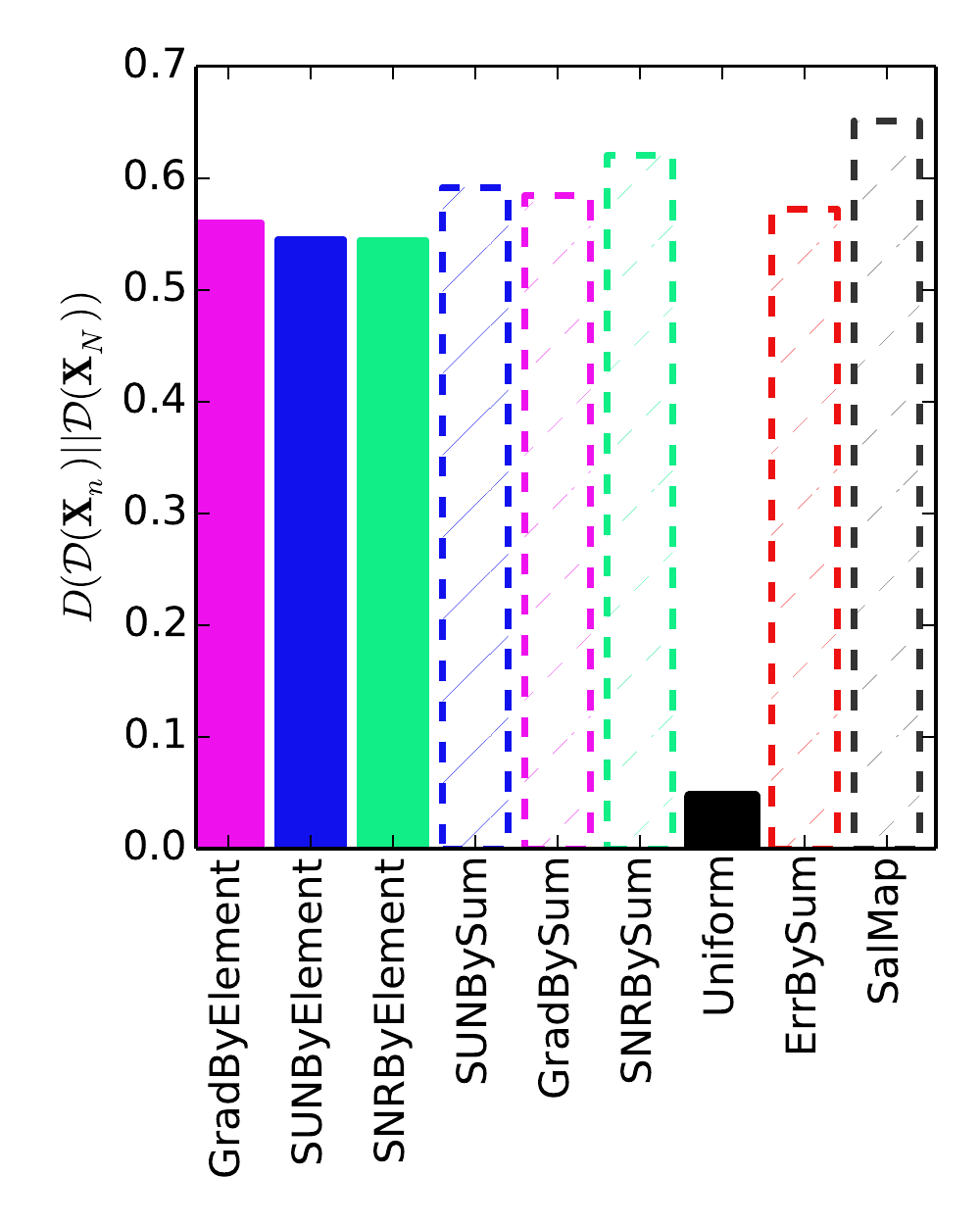}} \\

\multicolumn{2}{c}{(a) Gabor dictionary} & \multicolumn{2}{c}{(b) Alphanumeric dictionary} \\

\end{tabular}
\caption{Characterization of ${\bf X}_n$.  Left columns: SNR of ${\bf X}_n$ (higher is better).  Right columns: $D(\mathcal{D}({\bf X}_n) || \mathcal{D}({\bf X}_N))$ (lower is better).}
\label{fig:snr}
\end{figure*}

There are several directions to which we plan to extend this work.  One is the theoretical analysis of the selection algorithms. For instance, we did not explore under what conditions learning with example selection leads to the same solutions as an unbiased learning, although empirically we observed that to be the case.  As in the curriculum learning paradigm~\cite{bengio2009curriculum}, it is also possible that different selection algorithms are better suited at different stages of learning.  Another is to apply the active example selection processes to hierarchical architectures such as stacked autoencoders and Restricted Boltzmann Machines.  In these cases, an interesting question arises as to how information from each layer should be combined to make the selection decision.  We intend to explore some of these questions in the future using learning tasks similar to this work.

\bibliography{ExampleSelection}
\bibliographystyle{plainnat}

\end{document}